\documentclass{article}

\PassOptionsToPackage{numbers, compress}{natbib}

\usepackage[preprint]{neurips_2019}

\usepackage[usenames,dvipsnames]{xcolor}
\usepackage[utf8]{inputenc}
\usepackage[T1]{fontenc}
\usepackage[colorlinks, linktoc=all]{hyperref}
\usepackage[all]{hypcap}
\hypersetup{citecolor=Violet}
\hypersetup{linkcolor=black}
\hypersetup{urlcolor=MidnightBlue}
\usepackage{url} 
\usepackage{booktabs} 
\usepackage{amsmath,amsthm,amssymb,amsfonts}
\usepackage{nicefrac}  
\usepackage{microtype}
\usepackage{comment}
\usepackage{natbib}
\usepackage{algorithm}
\usepackage{algorithmic}

\usepackage{bm}
\usepackage{enumerate}
\usepackage[export]{adjustbox}
\usepackage[nameinlink]{cleveref}

\usepackage{graphicx}
\usepackage{subfigure}

\crefname{equation}{eq.}{eqs.}
\Crefname{equation}{Eq.}{Eqs.}
\creflabelformat{equation}{#1#2#3}

\newtheorem{proposition}{Proposition}

\newtheorem{definition}{Definition}

\usepackage[acronym,smallcaps,nowarn]{glossaries}
\newacronym{VAE}{vae}{Variational Auto-Encoder}
\newacronym{CVAE}{cvae}{Correlated Variational Auto-Encoder}
\newacronym{ACVAE}{acvae}{Adaptive Correlated Variational Auto-Encoder}

\title{Learning Correlated Latent Representations with Adaptive Priors}

\author{%
  Da Tang\\
  Columbia University\\
  \texttt{datang@cs.columbia.edu} \\
   \And
  Dawen Liang \\
  Netflix Inc. \\
  \texttt{dliang@netflix.com} \\
  \AND
  Nicholas Ruozzi \\
  The University of Texas at Dallas \\
  \texttt{nicholas.ruozzi@utdallas.edu} \\
  \And
  Tony Jebara \\
  Columbia University \& Spotify Inc.\\
  \texttt{jebara@cs.columbia.edu} \\
}

\begin{document}

\maketitle

\begin{abstract}
\Acrlongpl{VAE} (\acrshortpl{VAE}) have been widely applied for learning compact, low-dimensional latent representations of high-dimensional data. When the correlation structure among data points is available, previous work proposed \acrlongpl{CVAE} (\acrshortpl{CVAE}), which employ a structured mixture model as prior and a structured variational posterior for each mixture component to enforce that the learned latent representations follow the same correlation structure. However, as we demonstrate in this work, such a choice cannot guarantee that \acrshortpl{CVAE} capture all the correlations.  Furthermore, it prevents us from obtaining a tractable joint and marginal variational distribution. To address these issues, we propose \acrlongpl{ACVAE} (\acrshortpl{ACVAE}), which apply an adaptive prior distribution that can be adjusted during training and can learn a tractable joint variational distribution. Its tractable form also enables further refinement with belief propagation. Experimental results on link prediction and hierarchical clustering show that \acrshortpl{ACVAE} significantly outperform \acrshortpl{CVAE} among other benchmarks.
\end{abstract}

\section{INTRODUCTION}
\Acrlongpl{VAE} (\acrshortpl{VAE}) \citep{kingma2013auto,rezende2014stochastic} are a family of deep generative models that learn latent embeddings for data. By applying variational inference on the latent variables, \acrshortpl{VAE} learn a stochastic mapping from high-dimensional data to low-dimensional representations, which can be used for many downstream tasks, including classification, regression, and clustering.

\acrshortpl{VAE} assume the data points are $i.i.d.$ generated and treat the model and posterior approximations as factorized over data points. However, if we know \emph{a priori} that there is structured correlation between the data points, e.g., for graph-structured datasets \citep{shi2014correlated,bruna2013spectral,hamilton2017inductive,tang2019correlated}, correlated variational approximations can help. \citet{tang2019correlated} proposed \acrlongpl{CVAE} (\acrshortpl{CVAE}), which take this kind of correlation structure as auxiliary information to guide the variational approximations for the latent embeddings by constructing a prior from a uniform mixture of tractable distributions on \emph{maximal acyclic subgraphs} of the given undirected correlation graph. 

However, there are several limitations that potentially prevent \acrshortpl{CVAE} from learning better correlated latent embeddings. First, it is possible that some of the maximal acyclic subgraphs of the given graph can, by themselves, well-capture the correlation between the data points while others may poorly capture the correlation. As a result, taking a uniform average may yield a sub-optimal result. Second, while the prior in \acrshortpl{CVAE} is over multiple subgraphs, each subgraph has a unique joint variational distribution, and there is no single \emph{global} joint variational distribution over the latent variables. \acrshortpl{CVAE} do learn pairwise variational approximation functions, but they are \textbf{not} exact pairwise marginal variational distributions on the latent variables. As a result, applying these variational approximation functions to some downstream tasks, e.g. link prediction, may result in poor performances due to the inexact approximations. In addition, \acrshortpl{CVAE} require a pre-processing step that takes an amount of time cubic in the number of vertices, which limits its applicability to smaller datasets.

To address these issues, we propose \acrlongpl{ACVAE} (\acrshortpl{ACVAE}), which chooses a non-uniform average over tractable distributions over the maximal acyclic subgraphs as a prior. This prior is adaptive, and will be adjusted during optimization. To learn the mixture weights, we provide two options, empirical Bayes or saddle-point optimization, both of which maximize the objective with respect to the model and variational parameters. The difference is that while empirical Bayes also maximizes the objective with respect to the prior structure, saddle-point optimization seeks to optimize the objective under the \emph{worst} prior for more robust inference. In both cases, the non-uniform average converges to a tractable prior on a single graph, which ensures that we obtain a holistic tractable joint variational distribution. With this variational distribution, we obtain exact marginal evaluation using exact inference algorithms, e.g., belief propagation. Moreover, \acrshortpl{ACVAE} do not require the cubic time pre-processing step embedded in \acrshortpl{CVAE}, and they are generally faster for evaluation in practice. We demonstrate the superior empirical performance of \acrshortpl{ACVAE} for link prediction and hierarchical clustering on various real datasets.

\section{VAES WITH CORRELATIONS}
In this section, we provide a brief overview of \acrlongpl{VAE} (\acrshortpl{VAE}) \citep{kingma2013auto,rezende2014stochastic} as well as \acrlongpl{CVAE} (\acrshortpl{CVAE}) \citep{tang2019correlated}, which take the correlation structure among data points into consideration. 

\subsection{Variational Auto-Encoders}
We use a latent variable model to fit data $\bm x=\{\bm x_1,\ldots,\bm x_n\}\subset\mathbb R^D$. The model assumes that there exist low-dimensional latent embeddings for each data point $\bm z=\{\bm z_1,\ldots,\bm z_n\}\subset\mathbb R^d$ ($d\ll D$), which come from a prior distribution $p_0(\cdot)$, and $\bm x_i$'s are drawn conditionally independently given $\bm z_i$. Denote the model parameters as $\bm\theta$. The likelihood of this model is $p_{\bm\theta}(\bm x)=\prod\limits_{i=1}^n\int p_0(\bm z_i)p_{\bm\theta}(\bm x_i | \bm z_i)d\bm z_i$.

To simultaneously learn the model parameters $\bm\theta$ as well as a mapping from the observed data $\bm x$ to the latent embeddings $\bm z$, \glspl{VAE} \citep{kingma2013auto,rezende2014stochastic} apply a data-dependent variational approximation $q_{\bm\lambda}(\bm z | \bm x)=\prod\limits_{i=1}^n q_{\bm\lambda}(\bm z_i |\bm x_i)$, where $\bm\lambda$ denotes the variational parameters, and maximize the \textit{evidence lower-bound} (ELBO) on the log-likelihood of the data, $\log p_{\bm\theta}(\bm x)$:
\begin{align}
L(\bm\lambda,\bm\theta) &=\mathbb E_{q_{\bm\lambda}(\bm z | \bm x)}\left[\log p_{\bm\theta}(\bm x | \bm z) \right]-\text{KL}(q_{\bm\lambda}(\bm z | \bm x)||p_0(\bm z)).
\label{eqn:elbo}
\end{align}

\subsection{\acrlongpl{CVAE}} \label{sec:cvae}
Standard \glspl{VAE} are capable of learning compact low-dimensional embeddings for high-dimensional data. 
However, due to the i.i.d. assumption, they fail to account for the correlations between data points when \emph{a priori} we know such correlations exist. 
\glspl{CVAE} \citep{tang2019correlated} mitigate the issue by employing a structured prior as well as a structured variational posterior.

Formally, assume we are given an undirected correlation graph $G=(V, E)$, where $(v_i, v_j)\in E$ represents that the data points $x_i$ and $x_j$ are correlated. \glspl{CVAE} apply a correlated prior $p^\textrm{corr}_0(\bm z)$ on the latent variables $\bm z_i$'s
which satisfies
\begin{equation}
\begin{cases}
p_0^{\textrm{corr}}(\bm z_i)=p_0(\bm z_i)\text{ for all }v_i\in V\\
p_0^{\textrm{corr}}(\bm z_i,\bm z_j)=p_0(\bm z_i, \bm z_j)\text{ if }(v_i, v_j)\in E.
\end{cases}
\label{eqn:prior-pair-acyclic}
\end{equation}
Here $p_0(\cdot)$ and $p_0(\cdot,\cdot)$ are parameter-free functions that capture the singleton and pairwise marginal distributions of the latent variables. 
For example, we can set $p_0(\cdot)$ to be the density of a standard multivariate normal distribution and $p_0(\cdot,\cdot)$ to be a multivariate normal density that has high values if the two inputs are close to each other. 
With such a prior, we again assume $\bm x_i$'s are drawn conditionally independently given $\bm{z}_i$.
When $G$ is acyclic, such a prior $p_0^{\text{corr}}(\bm z)$ does exist \citep{wainwright2008graphical}:
\begin{equation}
p_0^{\textrm{corr}}(\bm z)=\prod\limits_{i=1}^np_0(\bm z_i)\prod\limits_{(v_i,v_j)\in E}\frac{p_0(\bm z_i,\bm z_j)}{p_0(\bm z_i)p_0(\bm z_j)}.
\label{eqn:prior-acyclic}
\end{equation}
However, when $G$ is not acyclic, \Cref{eqn:prior-acyclic} is not necessarily a valid probability density function. 
To deal with this issue, \glspl{CVAE} propose constructing a prior that is a mixture over the set $\mathcal{A}_G$ of all of $G$'s \textit{maximal acyclic subgraphs}, which are defined as follows.
\begin{definition}[Maximal acyclic subgraph]
For an undirected graph $G=(V,E)$, an acyclic subgraph $G'=(V', E')$ is a maximal acyclic subgraph of $G$ if:
\begin{itemize}
\item $V'=V$, i.e., $G'$ contains all vertices of $G$.
\item Adding any edge from $E/E'$ to $E'$ will create a cycle in $G'$.
\end{itemize}
\end{definition}
Each of the maximal acyclic subgraphs $G' \in \mathcal{A}_G$ partially approximates the correlation structure in $G$, and \glspl{CVAE} set the prior $p_0^{\textrm{corr}_g}(\bm z)$ to be the uniform average over all of these tractable densities.
 \begin{equation}
p_0^{\textrm{corr}_g}(\bm z)\triangleq \frac{1}{|\mathcal A_G|}\sum\limits_{G'=(V,E')\in\mathcal A_G} p_0^{G'}(\bm z),
\label{eqn:prior}
\end{equation}
where $p_0^{G'}(\bm z)=\prod\limits_{i=1}^np_0(\bm z_i)\prod\limits_{(v_i,v_j)\in E'}\frac{p_0(\bm z_i,\bm z_j)}{p_0(\bm z_i)p_0(\bm z_j)}$ is a prior on a maximal acyclic subgraph $G'=(V,E')$ with the same form as in \Cref{eqn:prior-acyclic}. For each $G' \in \mathcal{A}_G$, we can similarly define a structured variational approximation $q_{\bm\lambda}^{G'}(\bm z | \bm x)$ following the form of \Cref{eqn:prior-acyclic} (see \Cref{app:cvae} for details). With this structured prior and variational posterior, \glspl{CVAE} optimize a different ELBO:
\begin{equation}
\begin{aligned}
\log p_{\bm\theta}(\bm x)&=\log\mathbb E_{p_0^{\textrm{corr}_g}(\bm z)}[p_{\bm\theta}(\bm x | \bm z)]\\
&\ge\frac1{|\mathcal A_G|}\sum\limits_{G'\in\mathcal A_G}\Big(\mathbb E_{q_{\bm\lambda}^{G'}(\bm z | \bm x)}[\log p_{\bm\theta}(\bm x | \bm z)] -\text{KL}(q_{\bm\lambda}^{G'}(\bm z | \bm x) || p_0^{G'}(\bm z))\Big).\label{eqn:lb1}
\end{aligned}
\end{equation}
Even though empirically \citet{tang2019correlated} show that \glspl{CVAE} are capable of capitalizing on the correlation structure as auxiliary information when learning latent embeddings. As discussed in the introduction, there are a few limitations to this approach. 
In the next section, we will propose fixes to all of these limitations.

\section{ADAPTIVE CORRELATED VAES}
\label{sec:acvae}
\subsection{A Non-uniform Mixture Prior}
As motivated in \Cref{sec:cvae}, rather than using a uniform average, we instead employ a categorical distribution $\bm\pi\in\triangle^{|\mathcal A_G|-1}$ representing the normalized weights over all maximal acyclic subgraphs $G'\in\mathcal A_G$ of $G$.
In the ELBO in \Cref{eqn:lb1}, we can replace the uniform average in the prior $p_0^{\textrm{corr}_g}(\bm z)$ in \Cref{eqn:prior} with the non-uniform distribution $\bm\pi$, which gives us the following ELBO:
\begin{equation}
\begin{aligned}
&\mathbb E_{G' \sim \bm\pi}\left[\mathbb E_{q_{\bm\lambda}^{G'}(\bm z | \bm x)}[\log p_{\bm\theta}(\bm x | \bm z)]-\text{KL}(q_{\bm\lambda}^{G'}(\bm z | \bm x) || p_0^{G'}(\bm z))\right]\\
&\le\mathbb E_{G' \sim \bm\pi}\left[\mathbb E_{p_0^{G'}(\bm z)}[\log p_{\bm\theta}(\bm x | \bm z)]\right]:=\mathbb E_{p_0^{\bm\pi}(\bm z)}[\log p_{\bm\theta}(\bm x | \bm z)]\\
&\le\log p_{\bm\pi,\bm\theta}(\bm x).
\end{aligned}
\label{eqn:lb2}
\end{equation}
Here we define the non-uniform prior $p_0^{\bm\pi}(\bm z) = \mathbb{E}_{G' \sim \bm\pi}[p_0^{G'}(\bm z)]$. From the above inequality we can see that, using the non-uniform prior $p_0^{\bm\pi}$, we are still able to obtain a lower bound of the log-likelihood $\log p_{\bm\pi,\bm\theta}(\bm x)$, which is now also parametrized by the weight parameter $\bm\pi$. 
If we optimize $\bm\pi$ together with all the other parameters, the above loss function implies that we are optimizing with an adaptive prior. Hence, we call the above model \glspl{ACVAE}. If we replace $\bm\pi$ with a uniform distribution over all subgraphs in $\mathcal A_G$, we recover \glspl{CVAE}.

Plugging $q_{\bm\lambda}^{G'}(\bm z | \bm x)$ and $p_0^{G'}(\bm z)$ from \Cref{sec:cvae} into \Cref{eqn:lb2}, yields the following ELBO for \glspl{ACVAE}:
\begin{equation}
\begin{aligned}
&\mathcal L^{\textrm{ACVAE}}(\bm\pi, \bm\lambda, \bm\theta):=\sum\limits_{i=1}^n\Big(\mathbb E_{q_{\bm\lambda}(\bm z_i | \bm x_i)}\left[\log p_{\bm\theta}(\bm x_i | \bm z_i) \right] -\text{KL}(q_{\bm\lambda}(\bm z_i | \bm x_i)||p_0(\bm z_i))\Big)-\sum\limits_{(v_i,v_j)\in E}w^{\text{MAS}}_{G, \bm\pi, (v_i, v_j)}\cdot\\
&\Big(\text{KL}(q_{\bm\lambda}(\bm z_i, \bm z_j | \bm x_i, \bm x_j)||p_0(\bm z_i, \bm z_j))-\text{KL}(q_{\bm\lambda}(\bm z_i | \bm x_i)||p_0(\bm z_i)) - \text{KL}(q_{\bm\lambda}(\bm z_j | \bm x_j)||p_0(\bm z_j))\Big).
\end{aligned}
\label{eqn:loss-acvae-nons}
\end{equation}
Similar to \glspl{CVAE}, we have edge weights
$w^{\text{MAS}}_{G, \bm\pi, (v_i, v_j)}$ representing the expected appearance probability for edge $(v_i, v_j)$ over the set of maximal acyclic subgraphs $\mathcal{A}_G$ given the distribution $\bm\pi$. In the following definition, we abusively write $\bm\pi(G')$ as the probability of $G'$ being sampled from $\mathcal{A}_G$.
\begin{definition}[Non-uniform maximal acyclic subgraph edge weight]
For an undirected graph $G=(V, E)$, an edge $e\in E$ and a distribution $\bm\pi$ on the set $\mathcal A_G$ of maximal acyclic subgraphs  of $G$, define $w^{\text{MAS}}_{G, \bm\pi, e}$ to be the expected appearance probability of the edge $e$ in a random maximal acyclic subgraph $G'=(V,E') \sim \bm\pi$, i.e., $w^{\text{MAS}}_{G, \bm\pi, e}:=\sum\limits_{G'\in\mathcal A_G, e\in E'}\bm\pi(G')$.  
\label{def:numasew}
\end{definition}

Similar to \acrshortpl{CVAE},  we can apply negative sampling (equivalent to applying a complete graph as a weak prior) to \acrshortpl{ACVAE} as regularization, which helps prevent overfitting on the learned pairwise variational approximation ($\gamma>0$ is the regularization strength):
\begin{equation}
\begin{aligned}
\mathcal L^{\textrm{ACVAE}\textrm{-NS}}(\bm\pi, \bm\lambda, \bm\theta):=&\mathcal L^{\textrm{ACVAE}}(\bm\pi, \bm\lambda, \bm\theta)-\gamma\cdot \Big(\sum\limits_{i=1}^n\text{KL}(q_{\bm\lambda}(\bm z_i | \bm x_i)||p_0(\bm z_i))\\
&+\frac{2}n\sum\limits_{1\le i < j\le n}\mathbb E_{q_{\bm\lambda}(\bm z_i, \bm z_j | \bm x_i, \bm x_j)}\log\frac{q_{\bm\lambda}(\bm z_i, \bm z_j | \bm x_i,\bm x_j)}{q_{\bm\lambda}(\bm z_i | \bm x_i)q_{\bm\lambda}(\bm z_j | \bm x_j)}\Big). 
\end{aligned}
\label{eqn:loss-acvae}
\end{equation}
In what follows, we use $\mathcal{L}^{\textrm{ACVAE}}$ to refer to $\mathcal{L}^{\textrm{ACVAE}\textrm{-NS}}$ for notational brevity.

\subsection{Learning the Non-uniform Mixture} \label{sec:minimax}
With the loss function in \Cref{eqn:loss-acvae}, an intuitive direction for estimating $\bm\pi$ would be to perform empirical Bayes \citep{efron2012large} and directly maximize $\mathcal L^{\textrm{ACVAE}}(\bm\pi, \bm\lambda, \bm\theta)$ with respect to $\bm\pi$, $\bm\lambda$ and $\bm\theta$, as in \Cref{eqn:opt-empirical}: 
\begin{equation}
\max\limits_{\bm\lambda,\bm\theta}\max\limits_{\bm\pi}\mathcal L^{\textrm{ACVAE}}(\bm\pi,\bm\lambda, \bm\theta).\label{eqn:opt-empirical}
\end{equation}
Alternatively, we can consider a minimax saddle-point optimization, which may\textbf{} lead to more robust inference: 
\begin{equation}
\max\limits_{\bm\lambda,\bm\theta}\min\limits_{\bm\pi}\mathcal L^{\textrm{ACVAE}}(\bm\pi,\bm\lambda, \bm\theta).\label{eqn:opt-saddle}
\end{equation}
As \Cref{eqn:opt-saddle} indicates, we are optimizing the ELBO under the prior that produces the \textit{lowest} lower bound. 
The intuition is that if we can even optimize the worst lower bound well, the variational distribution and the model distribution we learn would be robust and generalize better. 
This is similar to the \emph{least favorable prior}, under which a Bayes estimator can achieve minimax risk \citep{lehmann2006theory}. 

Empirical Bayes (\Cref{eqn:opt-empirical}) aims to find the best variational approximation, while the saddle-point option (\Cref{eqn:opt-saddle}) aims for robust inference. At first glance, the empirical Bayes option seems more reasonable since it gives us the tightest lower bound. However, a better ELBO does not necessarily translate into better predictive performance in the downstream task. 
In \Cref{sec:expe}, we compare these two optimization options on various datasets, and discuss the pros and cons of each.

An important observation is that, no matter which option is applied, for fixed $\bm\lambda$ and $\bm\theta$, the loss function $\mathcal L^{\textrm{ACVAE}}$ is linear w.r.t. the weight parameter $\bm\pi$. Therefore, if optima for $\mathcal L^{\textrm{ACVAE}}(\bm\pi,\bm\lambda, \bm\theta)$ exist, then at least one optimum will have a $\bm\pi^*$ which puts all of its probability mass on a single subgraph $G'^*$. 
\begin{proposition}[Optimum for $\bm\pi$]\label{prop:optimal}
If the optimization in \Cref{eqn:opt-empirical} or \Cref{eqn:opt-saddle} has global optima, then at least one optimum $(\bm\pi^*,\bm\lambda^*,\bm\theta^*)$ will have a $\bm\pi^*$  that places all of its probability mass on a single maximal acyclic subgraph $G'^*\in\mathcal A_G$.
\end{proposition}
From this proposition, we know both \Cref{eqn:opt-empirical} and \Cref{eqn:opt-saddle} return a single subgraph $G'^*$, which drastically simplifies the structured prior. At this optimum, the loss function becomes the ELBO on a single acyclic subgraph $G'^*$, with $q_{\bm\lambda^*}^{G'^*}(\bm z | \bm x)$ as the variational distribution. Therefore, we have a holistic variational approximation, overcoming a limitations of \glspl{CVAE}.

\subsection{Learning with Alternating Updates}
Direct optimization of either \Cref{eqn:opt-empirical} or \Cref{eqn:opt-saddle} is non-trivial. Following similar saddle-point optimization for a spanning tree structured upper bound for the log-partition function of undirected graphical models \citep{wainwright2002stochastic,wainwright2005new}, we perform an alternating optimization procedure on the parameters $\bm\lambda$, $\bm\theta$ and $\bm\pi$. Details are shown in \Cref{alg:updates}.

\paragraph{Updates For $\bm\pi$}
When the parameters $\bm\lambda$ and $\bm\theta$ are fixed, the loss function $\mathcal L^{\textrm{ACVAE}}(\bm\pi,\bm\lambda, \bm\theta)$ is linear in $\bm\pi$. However, we cannot directly optimize over $\bm\pi\in\triangle^{|\mathcal A_G|-1}$, as it may contain exponentially many dimensions. We can instead update the edge weights $w^{\text{MAS}}_{G, \bm\pi, (v_i, v_j)}$ as the loss function is also linear in them.

By definition, we know that each maximal acyclic subgraph $G'$ of $G$ is a forest, consisting of one spanning tree for each connected component of $G$. Therefore, the domain for the edge weights $\bigcup\limits_{e\in E}\{w^{\text{MAS}}_{G, \bm\pi, e}\}$ is the projection of the Cartesian product of the spanning tree polytopes for all connected components of $G$ \citep{wainwright2002stochastic,wainwright2005new} to the edge weight space. This Cartesian product on the polytopes is convex and its boundary is determined by potentially exponentially many linear inequalities. Despite that, directly maximizing (or minimizing) $\mathcal L^{\textrm{ACVAE}}(\bm\pi,\bm\lambda, \bm\theta)$ with respect to these weights $\bigcup\limits_{e\in E}\{w^{\text{MAS}}_{G, \bm\pi, e}\}$ is in fact tractable: the optimum for \Cref{eqn:opt-empirical} or \Cref{eqn:opt-saddle} is obtained at $\hat{\bm\pi}$ that has all the mass on a single maximal acyclic subgraph $\hat G'$. This means the optimum for these edges weights can be obtained from a single subgraph $\hat G'$. By re-arranging terms in \Cref{eqn:loss-acvae} with respect to $\bigcup\limits_{e\in E}\{w^{\text{MAS}}_{G, \bm\pi, e}\}$, it is not difficult to see that $\hat G'$ should have the smallest (for empirical Bayes) or largest (for saddle-point) ``edge mass'' sum over all maximal acyclic subgraphs $\mathcal A_G$, where the ``edge mass'' $m_{(v_i,v_j)}$ of edge $e=(v_i,v_j)$ is:
\begin{equation}
\begin{aligned}
m_{(v_i,v_j)}:=\text{KL}(q_{\bm\lambda}(\bm z_i, \bm z_j | \bm x_i, \bm x_j)||p_0(\bm z_i, \bm z_j))-\text{KL}(q_{\bm\lambda}(\bm z_i | \bm x_i)||p_0(\bm z_i)) - \text{KL}(q_{\bm\lambda}(\bm z_j | \bm x_j)||p_0(\bm z_j)),
\label{eqn:mass}
\end{aligned}
\end{equation}
which means $\hat{G}'$ is the combination of the minimum (for empirical Bayes) or maximum (for saddle-point) spanning trees of all connected components of the graph with $m_{(v_i, v_j)}$ as the weights. 

Once we identify $\hat{G}'$, the optimal weights $\hat w^{\text{MAS}}_{G, \bm{\hat\pi}, e}$ are either 1 (if the edge $e$ is selected) or 0 (otherwise). 
Instead of directly updating the weights to the optimal values, we perform a soft update with step size $\alpha^t$ at iteration $t$, similar to \citet{wainwright2002stochastic,wainwright2005new}:
\begin{equation}
w^{\text{MAS}^{t+1}}_{G, \bm\pi, e}\leftarrow (1-\alpha^t)w^{\text{MAS}^t}_{G, \bm\pi, e}+\alpha^t\hat w^{\text{MAS}}_{G, \bm{\hat\pi}, e}.
\label{eqn:soft-update}
\end{equation}
This soft update helps prevent the algorithm from becoming trapped in bad local optima early in the optimization procedure. The step size $\alpha^t$ can be either a constant or dynamically adjusted during optimization. We set it to be a constant in our experiments. 

One of the limitations of \glspl{CVAE} mentioned in \Cref{sec:cvae} is the $O(|V|^3)$ pre-processing step to compute all the edge weights $w^{\text{MAS}}_{G, e}$. We alleviate this bottleneck in \glspl{ACVAE}, as it only takes $O(\min(|V|^2,|E|\log|V|))$ operations per initialization (details in \Cref{app:exp_proto}) and per update on the weights, which ensures that \glspl{ACVAE} can scale to datasets with many more vertices than would be feasible with \glspl{CVAE}. 
\paragraph{Updates For $\bm\lambda$ And $\bm\theta$}
When $\bm\pi$ is fixed, $\bm\lambda$ and $\bm\theta$ can be updated by taking a stochastic gradient step following $\nabla_{\bm\lambda,\bm\theta}\mathcal L^{\textrm{ACVAE}}(\bm\pi,\bm\lambda, \bm\theta)$ with reparametrization gradient \citep{kingma2013auto,rezende2014stochastic}, as done in standard \glspl{VAE}.
\begin{algorithm}[ht]
   \caption{\glspl{ACVAE} learning}
   \label{alg:updates}
\begin{algorithmic}
   \STATE {\bfseries Input:} data $\bm x_1,\ldots,\bm x_n\in\mathbb R^D$, undirected graph $G=(V=\{v_1,\ldots,v_n\}, E)$, parameter $\gamma>0$.
   \STATE Initialize the parameters $\bm\lambda$, $\bm\theta$. Initialize the weights $w^{\text{MAS}}_{G, \bm\pi, e}$ for each $e\in E$.
   \WHILE {not converged}
   \STATE Optimize the parameters $(\bm\lambda, \bm\theta)$ using the gradients $\nabla_{\bm\lambda,\bm\theta}\mathcal L^{\textrm{ACVAE}}(\bm\pi,\bm\lambda, \bm\theta)$.
   \STATE Compute the mass $m_{(v_i,v_j)}$ for each edge $e\in E$ with \Cref{eqn:mass}.
   \STATE Compute a minimum (if applying the empirical Bayes option in \Cref{eqn:opt-empirical}) or maximum (if applying the saddle-point option in \Cref{eqn:opt-saddle}) spanning tree of the graph with the masses $m_{(v_i,v_j)}$'s as the edge weights and update the weights $w^{\text{MAS}}_{G, \bm\pi, e}$ for each $e\in E$ according to \Cref{eqn:soft-update}.
   \ENDWHILE
   \STATE {\bfseries Return:} The parameters $\bm\lambda$, $\bm\theta$, the weights $w^{\text{MAS}}_{G, \bm\pi, e}$ for each $e\in E$.
\end{algorithmic}
\end{algorithm}

If empirical Bayes (\Cref{eqn:opt-empirical}) is applied, \Cref{alg:updates} will converge with properly selected learning rates. On the other hand, it is difficult to make any general statement about the convergence for saddle-point optimization (\Cref{eqn:opt-saddle}) since the objective is generally non-concave in $(\bm\lambda,\bm\theta)$. However, as we show in \Cref{sec:expe}, empirically we find that \Cref{alg:updates} is stable for both options and performs well on multiple real datasets.

\subsection{Exact Marginal Posterior Approximation with Belief Propagation}\label{sec:bp}
From \Cref{prop:optimal}, we know the weights $w^{\text{MAS}}_{G, \bm\pi, e}$ returned from \Cref{alg:updates} are from a single maximal acyclic subgraph $G' \in \mathcal{A}_G$. Consequently, we have a holistic variational approximation $q_{\bm\lambda}^{G'}(\bm z | \bm x)$. However, by itself this variational approximation might not be necessarily better at the downstream predictive tasks than \glspl{CVAE} since it can only make use of the structure from \emph{one} maximal acyclic subgraph $G'$. 

On the plus side, the acyclic structure of $G'$ makes it possible to compute the exact pairwise marginal variational distribution between any pair of vertices via a belief-propagation-style \citep{pearl1982reverend} message-passing algorithm, which is not possible for \glspl{CVAE}, as it does not have a single joint variational distribution on $\bm z$. This can be crucial in tasks in which we need an accurate pairwise marginal approximation, e.g., link prediction and hierarchical clustering. 

Consider any $v_i\neq v_j\in V$ that are in the same connected component of $G'$. Since $G'$ is acyclic there is a unique path from $v_i$ to $v_j$.  Denote it as $v_i=u^{i,j}_0\rightarrow u^{i,j}_1\rightarrow\ldots\rightarrow u^{i,j}_{k_{i,j}}=v_j$. The exact pairwise marginal $r_{\bm\lambda}(\bm z_i, \bm z_j | \bm x_i, \bm x_j)$ equals
\begin{equation*}
\begin{aligned}
\int \prod\limits_{l=0}^{k_{i,j}-1}q_{\bm\lambda}(\bm z_{u^{i,j}_l}, \bm z_{u^{i,j}_{l+1}} | \bm x_{u^{i,j}_l}, \bm x_{u^{i,j}_{l+1}})\prod\limits_{l=1}^{k_{i,j}-1}\frac{d\bm z_{u^{i,j}_l}}{ q_{\bm\lambda}(\bm z_{u^{i,j}_l} | \bm x_{u^{i,j}_l})}.
\end{aligned}
\end{equation*}
The above pairwise marginal densities can be computed for all pairs of $(v_i, v_j)$ by doing a depth- or breadth-first search starting from each $v_i \in V$ after we obtain the variational approximation $q_{\bm\lambda}^{G'}(\bm z | \bm x)$ from \Cref{alg:updates}, which has a total complexity of $O(|V|^2)$. Note that the time complexity for evaluating every pairwise marginal in \glspl{CVAE} is also $O(|V|^2)$. But the belief propagation refinement computation is usually more efficient in practice, since it involves much less neural network function evaluations, which dominate the runtime.

\section{RELATED WORK}
This work extends \acrshortpl{CVAE} with the idea of learning a non-uniform average loss over some tractable loss functions on maximal acyclic subgraphs of the given graph. This is similar to the idea of obtaining a tighter upper bound on the log-partition function for an undirected graphical model by minimizing over a convex combination of spanning trees of the given graph \citep{wainwright2005new}.  
To optimize the parameters, \citet{wainwright2005new} also apply alternating updates on the parameters and the distributions over the spanning tress, similar to the approach in \acrshort{ACVAE} learning. Alternating parameter updates are useful for many other cases. For example, Alternating Least Squares for matrix factorization \citep{takane1977nonmetric} and Alternating Direction Method of Multipliers (ADMM) for convex optimization \citep{chambolle2011first,tang2015duality,hong2017linear}.

Some recent work also focuses on incorporating correlation structures over latent variables. For example, \citet{hoffman2015stochastic} proposed structured variational families that can improve over traditional mean-field variational inference. \citet{johnson2016composing} proposed Structured \acrshortpl{VAE} that apply more complex forms for the priors on the latent embeddings. Recently in the NLP community, \citet{yin2018structvae} proposed utilizing tree-structured latent variable models to deal with semantic parsing. However, most of these works focus on correlations \emph{within} dimensions of latent variables whereas our work focus on correlations \emph{between} latent variables, similar to the setting of \acrshortpl{CVAE}. In addition, \citet{luo2018semi} incorporated pairwise correlations between latent variables into deep generative models for semi-crowdsourced clustering.

Another line of related work appears in convolutional networks for graphs and their extensions \citep{bruna2013spectral,duvenaud2015convolutional,defferrard2016convolutional,niepert2016learning,hamilton2017inductive,velivckovic2018deep}, which also take graph structure of data into considerations.

\section{EXPERIMENTS}
\label{sec:expe}
In this section, we evaluate \glspl{ACVAE} on the task of link prediction and hierarchical clustering. We show that our method significantly outperforms various baselines. We attempt to identify the contributing factors for the gain, answering the following questions:

\textbf{Q1}: Uniform mixture (\acrshort{CVAE}) \emph{versus} non-uniform mixture (\acrshort{ACVAE}), which one is better? (\Cref{sec:expe-mixture})

\textbf{Q2}: How important is the belief propagation refinement for \acrshort{ACVAE}? (\Cref{sec:expe-mixture})

\textbf{Q3}: Empirical Bayes \emph{versus} saddle-point, which one performs better?
    Can we select purely based on ELBO? (\Cref{sec:expe-ebsp})
    
\textbf{Q4}: Does the learned single graph capture more information than singleton representations? What do the learned latent embeddings look like? (\Cref{sec:expe-vis})

\textbf{Q5}: Can \acrshort{ACVAE} scale to datasets that \acrshort{CVAE} cannot? (\Cref{sec:expe-scale})
\subsection{Experiment Settings}
Before presenting our experimental results, we describe the tasks, datasets, baslines, and metrics for evaluation. Additional details can be found in \Cref{app:exp}. 

\subsubsection{Tasks}
For each of the tasks, we are given a correlation graph $G=(V=\{v_1,\ldots,v_n\}, E)$ and a feature vector $\bm x_i\in\mathbb R^N$ for each $i\in\{1,\ldots,n\}$. 

For the link prediction task, we keep consistent with the setting of \citet{tang2019correlated}. 
For the hierarchical clustering experiments, we apply the \textbf{complete-linkage} algorithm \citep{yim2015hierarchical}, which is relatively more stable among common hierarchical clustering algorithms. We cluster all data points into $K=5$ clusters.

\subsubsection{Datasets}
We evaluate \glspl{ACVAE} on the following 3 datasets. All of 3 datasets are tested for link prediction and in addition the LibraryThing dataset is tested for the hierarchical clustering experiment:
\begin{itemize}
\item Epinions\footnote{\href{http://www.trustlet.org/downloaded\_epinions.html}{http://www.trustlet.org/downloaded\_epinions.html}} \citep{massa2007trust}, a public product rating dataset that contains $\approx49\text{K}$ users and $\approx140\text{K}$ products. After pre-processing, the dataset contains $\approx16{K}$ users.

\item Citation\footnote{\href{http://snap.stanford.edu/data/cit-HepTh.html}{http://snap.stanford.edu/data/cit-HepTh.html}} \citep{snapnets}, a High-energy physics theory citation network dataset, which has a citation graph with $\approx 28\text{K}$ papers and $\approx 353\text{K}$ citation edges. After preprocessing, the dataset contains $\approx2{K}$ users (for the results as in \Cref{sec:expe-mixture}). We also perform an experiment in \Cref{sec:expe-scale} on a larger version of this dataset, which contains $\approx26{K}$ users.

\item LibraryThing\footnote{\href{https://cseweb.ucsd.edu/~jmcauley/datasets.html\#social\_data}{https://cseweb.ucsd.edu/\textasciitilde jmcauley/datasets.html\#social\_data}} \citep{lakkaraju2013s}, a public book review data set that contains $\approx73\text{K}$ users and $\approx337\text{K}$ items. After pre-processing, the dataset contains $\approx6{K}$ users.
\end{itemize}

For the hierarchical clustering task, the LibraryThing dataset does not contain cluster labels for users. We generate the cluster labels for each user by learning a standard \acrshort{VAE} on the feature vectors $\bm x$, and perform the complete-linkage algorithm to cluster the data points into $K=5$ clusters. This helps us generate a \emph{semi-synthetic} dataset. 

\subsubsection{Baselines}
We compare \acrshort{ACVAE} with 4 baseline methods:
\begin{itemize}
\item \gls{VAE} \citep{kingma2013auto}: standard variational auto-encoders, with no information about the correlations.
\item GraphSAGE \citep{hamilton2017inductive}:  the state-of-the-art method for learning latent embeddings that takes the correlation structure into account with graph convolutional neural networks. 
\item {\sc cvae}$_\textrm{ind}$ and {\sc cvae}$_\textrm{corr}$ \citep{tang2019correlated}: Two variations of \glspl{CVAE} with factorized and structured variational approximations, respectively.
\end{itemize}
There are many different variants of GraphSAGE, and we applied one of them (details in \Cref{app:exp_proto}). It is possible that some other variants or parameter settings of this method may perform better on our tasks. But our main goal is not to derive a state-of-the-art method for these tasks. Instead, we aim to show insights on how to improve over standard \acrshortpl{VAE} and \acrshortpl{CVAE} through learning adaptive correlated priors.

\subsubsection{Metrics}
For all methods, we first learn latent embeddings $\bm z_1,\ldots,\bm z_n$, which are deterministic for GraphSAGE and stochastic for the \glspl{VAE}-based methods. Then we compute the pairwise distance $\text{dis}_{i,j}$ between each pair $(\bm z_i, \bm z_j)$ of the latent embeddings as $\Vert\bm z_i-\bm z_j\Vert_2^2$. Recall that the embeddings are stochastic for the \glspl{VAE}-based methods, hence we use $\mathbb{E}[\Vert\bm z_i-\bm z_j\Vert_2^2]$ as the pairwise distance. The expectation is taken over the variational pairwise marginal $q(\bm z_i, \bm z_j | \bm x_i, \bm x_j)$ or the refined pairwise marginal $r(\bm z_i, \bm z_j | \bm x_i, \bm x_j)$ if we perform belief propagation (\Cref{sec:bp}).

For the link prediction experiments, for each user $u_i$, we compute the \textit{Cumulative Reciprocal Rank} (CRR) as follows.
\[
\text{CRR}_i=\hspace{-.5cm}\sum\limits_{(v_i,v_j)\in E_{\text{test}}} \frac{1}{|\{k:(v_i,v_k)\not\in E_{\text{train}},\text{dis}_{i,k}\le\text{dis}_{i,j}\}|}.
\]
A larger CRR value indicates the heldout edges have a higher rank among all the candidates. We further normalize the CRR values to be in $[0,1]$, and report the normalized CRR (NCRR).

For hierarchical clustering, we apply the normalized mutual-information scores \citep{vinh2010information} as the metric. These scores are in the range $[0,1]$ and a larger score indicates better clustering performance.

\subsection{Results}

We show the heldout NCRR values for link predictions and the normalized MI scores in \Cref{tab:ncrr} and \Cref{tab:mi-score}, respectively. \acrshort{ACVAE}$_\textrm{EB}$ and \acrshort{ACVAE}$_\textrm{SP}$ stand for empirical Bayes (\Cref{eqn:opt-empirical}) and saddle-point optimization (\Cref{eqn:opt-saddle}), respectively. The rows with BP mean we perform belief-propagation refinement (\Cref{sec:bp}). We dissect the results in the following sections.  
\subsubsection{Advantages of the Non-uniform Mixture}
\label{sec:expe-mixture}
As motivated in \Cref{sec:cvae}, \glspl{ACVAE} improve over the limitations of \glspl{CVAE} by providing a holistic variational approximation at the end of the empirical Bayes or saddle-point optimization, which further enables applying belief propagation for more accurate marginal approximation. 

At first glance, the performance results in \Cref{tab:ncrr} for the single joint distribution (the rows \acrshort{ACVAE}$_\textrm{EB}$ and \acrshort{ACVAE}$_\textrm{SP}$) is no better than that of {\sc cvae}$_\textrm{corr}$, which applies a uniform mixture. We speculate in \Cref{sec:bp} that by itself this holistic variational approximation might not  necessarily be better at the downstream predictive tasks since it can only make use of the structure from \emph{one} maximal acyclic subgraph, even though it sometimes has a higher ELBO (\Cref{tab:elbo2}). However, we can observe a huge performance boost after applying the belief propagation refinement, which outperforms the baseline methods by a wide margin for link prediction and performs comparably better for hierarchical clustering.\footnote{\acrshort{VAE} does not count as a baseline method for the clustering experiment since it is applied as an oracle in the pre-processing steps. We omitted the results for \glspl{ACVAE} without belief propagation for the hierarchical clustering experiments since empirically we found their performance are much worse compared to the case of using belief propagation refinement.}

Recall that the prerequisite for applying the belief propagation is to have a variational distribution on a single acyclic subgraph (i.e., we \textbf{cannot} perform BP with \glspl{CVAE}).  
This answers two questions we sought to answer: First, the non-uniform mixture is not necessarily better than the uniform mixture at the downstream task even when it has a higher ELBO, but it opens up the possibility to perform exact inference; Second, variational approximations has a lot of room for improvement when compared with exact inference (i.e., belief propagation) on an acyclic graph. 
\begin{table}[t]
  \centering
  \caption{Link Prediction Normalized CRR}
\begin{tabular}{llll}
\\
    Method     & Epinions & Citation & LibraryThing      \\
    \hline
    \acrshort{VAE} &	$0.005\pm0.001$ & $0.018\pm0.004$ &  $0.006\pm0.000$   \\
    GraphSAGE &	$0.012\pm0.003$ & $0.020\pm0.002$ & $0.004\pm0.001$	\\
    {\sc cvae}$_\textrm{ind}$     &  $0.016\pm0.000$  & $0.040\pm0.003$  &  $0.012\pm0.001$  \\
    {\sc cvae}$_\textrm{corr}$     &     $0.017\pm0.001$  & $0.058\pm0.002$ &  $0.020\pm0.001$ \\
    \hline
    \acrshort{ACVAE}$_\textrm{EB}$ &	$0.013\pm0.000$ 	& $0.049\pm0.001$ &	$0.018\pm0.001$ \\
    \acrshort{ACVAE}$_\textrm{SP}$ &	 $0.010\pm0.003$	& $0.039\pm0.002$ &$0.018\pm0.001$ 	\\
    \acrshort{ACVAE}$_\textrm{EB+BP}$ & $0.034\pm0.003$	 	&  $\bm{0.126\pm0.005}$ & $\bm{0.032\pm0.002}$	 \\
    \acrshort{ACVAE}$_\textrm{SP+BP}$ & 	$\bm{0.035\pm0.001}$ 	& $0.123\pm0.007$ & $\bm{0.032\pm0.001}$ 	\\
  \end{tabular}
  \label{tab:ncrr}
\end{table}
\begin{table}[ht]
  \centering
  \caption{Hierarchical Clustering Normalized MI Scores}
\begin{tabular}{ll}
\\
    Method     & MI Scores     \\
    \hline
    GraphSAGE &	$0.002\pm0.000$	\\
    {\sc cvae}$_\textrm{ind}$ & $0.010\pm0.004$ \\
    {\sc cvae}$_\textrm{corr}$ & $0.002\pm0.000$ \\
    \acrshort{ACVAE}$_\textrm{EB+BP}$ &$\bm{0.012\pm0.003}$\\
    \acrshort{ACVAE}$_\textrm{SP+BP}$ &$0.011\pm0.002$ 	\\
  \end{tabular}
  \label{tab:mi-score}
\end{table}
\subsubsection{Empirical Bayes \emph{versus} Saddle-Point}
\label{sec:expe-ebsp}
As shown in \Cref{tab:ncrr} and \Cref{tab:mi-score}, both empirical Bayes and saddle-point optimization perform similarly on most tasks, though the saddle-point option is often more stable (normally having a smaller variance in the metrics). This is reasonable since the saddle-point objective optimizes the most conservative lower bound. 

Moreover, we show that we should not select between these two methods purely based on ELBO: By definition, the saddle-point optimization will yield an ELBO lower than empirical Bayes. In \Cref{tab:elbo} and \Cref{tab:elbo2} we report ELBO as well as NCRR for 4 choices of the negative sampling parameter $\gamma$ (\Cref{eqn:loss-acvae}) on Epinions and Citations with and without belief propagation refinement. We can see clearly that a better ELBO does not necessarily correlate with a better NCRR, regardless of belief propagation refinement or not.
\begin{table}[ht]
  \centering
  \caption{ELBO | Average NCRR Comparisons between ACVAE (with BP) on Epinions and Citation}
\begin{tabular}{lll}
\\
    Epinions     &  \acrshort{ACVAE}$_\textrm{EB+BP}$ & \acrshort{ACVAE}$_\textrm{SP+BP}$  \\
    \hline
    $\gamma=0.001$ & -31.8 | 0.034 & -31.9 | 0.031 \\
    $\gamma=0.1$ & -36.4 | 0.031 & -38.3 | 0.035 \\
    $\gamma=10.$ & -61.3 | 0.028 & -119 | 0.034 \\
    $\gamma=1000.$ & -674 | 0.028 & -1535 | 0.037\\
    \hline
    Citation     &  \acrshort{ACVAE}$_\textrm{EB+BP}$ &
    \acrshort{ACVAE}$_\textrm{SP+BP}$  \\
    \hline
    $\gamma=0.001$ & -7.48 | 0.126& -7.48 | 0.124 \\
    $\gamma=0.1$ & -7.91 | 0.113 & -8.59 | 0.121\\
    $\gamma=10.$ & -24.4 | 0.112 & -49.2 | 0.120\\
    $\gamma=1000.$ & -184 | 0.099& -288 | 0.054 \\
  \end{tabular}
  \label{tab:elbo}
\end{table}
\begin{table}[ht]
  \centering
  \caption{ELBO | Average NCRR Comparisons between ACVAE (without BP) and CVAE On Citation}
\begin{tabular}{llll}
\\
    Citation     &  \acrshort{ACVAE}$_\textrm{EB}$ & \acrshort{ACVAE}$_\textrm{SP}$ & \acrshort{CVAE} \\
    \hline
    $\gamma=0.001$ & -7.47 | 0.012 & -7.48 | 0.010 & -7.48 | 0.011\\
    $\gamma=0.1$ & -7.88 | 0.031 & -8.51 | 0.025 &  -8.49 | 0.023\\
    $\gamma=10.$ & -23.8 | 0.043 & -47.9 | 0.037 & -42.3 | 0.042\\
    $\gamma=1000.$ & -183 | 0.049& -286 | 0.039 & -267 | 0.058\\
  \end{tabular}
  \label{tab:elbo2}
\end{table}

In general, both methods have their advantages. On simpler datasets, e.g., Citation, on which all methods perform well, empirical Bayes is preferred since it can easily capture the best correlation structure. On the other hand, with more complex datasets/difficult tasks,  
saddle-point optimization tends to provide more robust inference and stable results. 

\subsubsection{Learned Graph Structures}
\label{sec:expe-vis}
\begin{figure}[ht]
  \centering
  \subfigure[Empirical Bayes]{
  \includegraphics[width=0.4\textwidth]{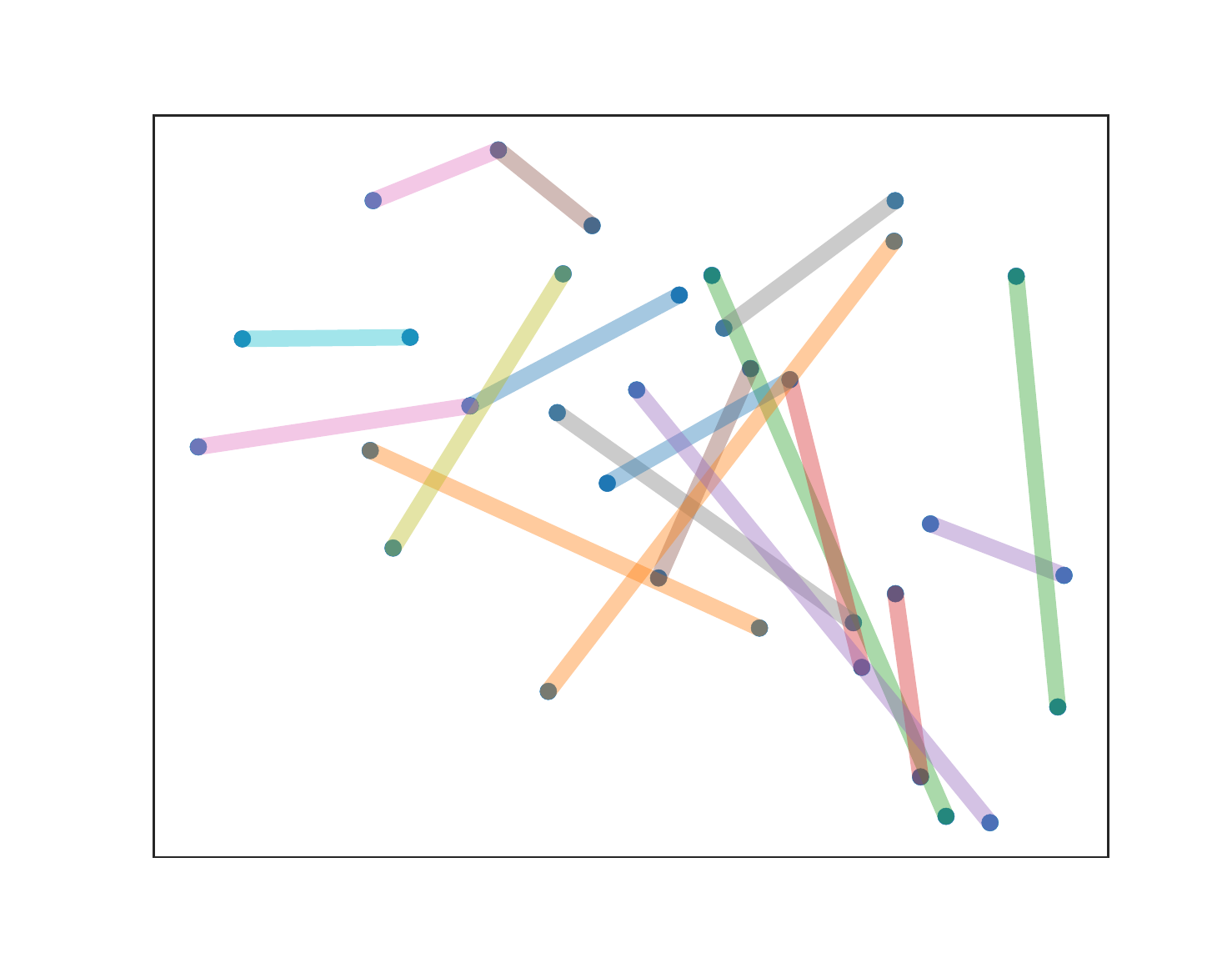}}
  \subfigure[Saddle-point]{
  \includegraphics[width=0.4\textwidth]{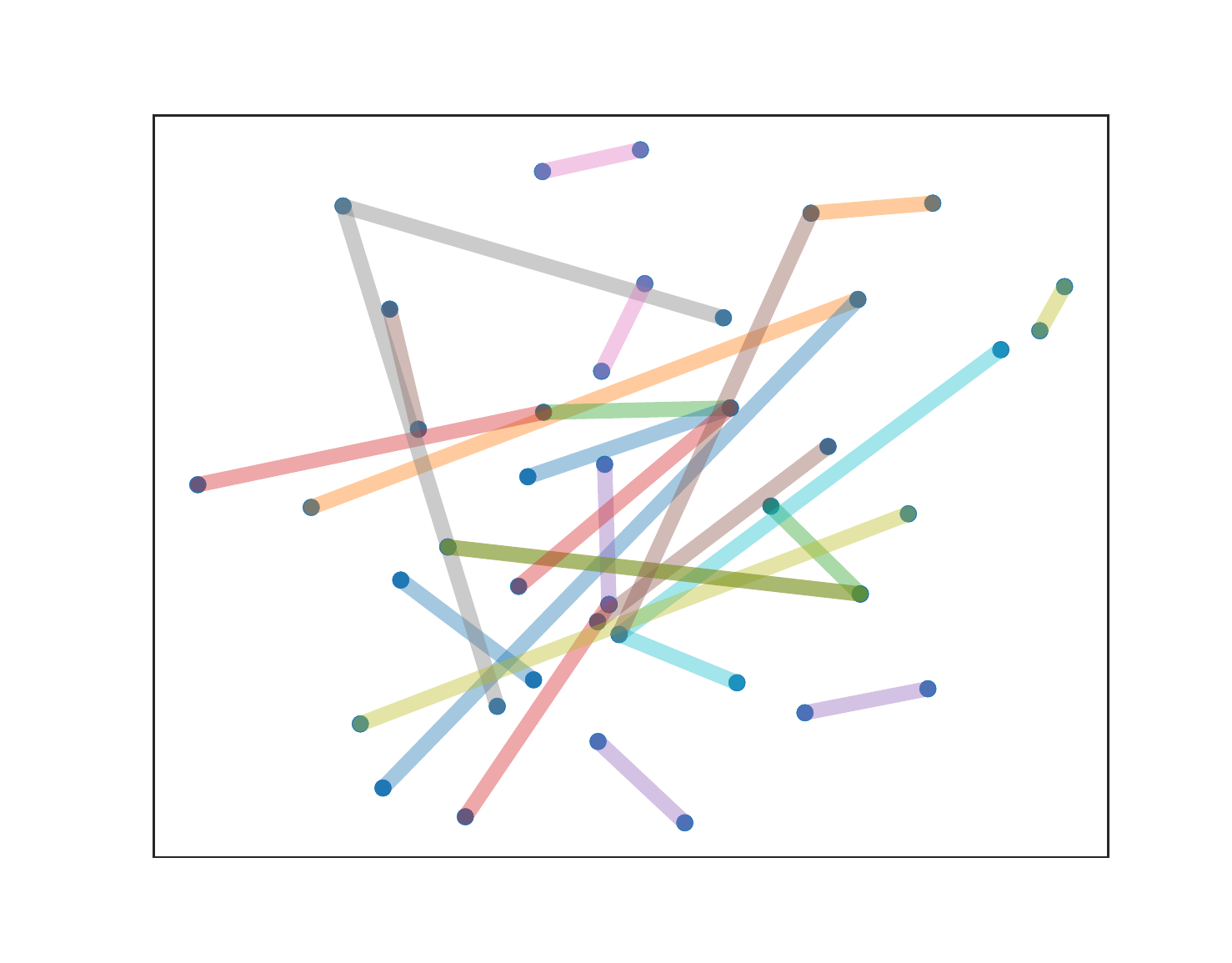}
  }
  \caption{Embeddings of the Learned Maximal Acyclic Subgraphs $\hat G'$}
   \label{fig:vis}
\end{figure}

In \Cref{fig:vis} we visualize part of the largest connected component of the maximal acyclic subgraph $\hat G'=(V,\hat E')$ that \acrshortpl{ACVAE} learn for the variational distribution on the Citation dataset with both empirical Bayes and saddle-point optimization (colors for better clarity only). The coordinates are t-SNE embeddings for the variational approximation mean of the latent variables. The edge widths are proportional to the strength of the learned correlations. We can see some of the learned embeddings are not necessarily close to each other even when they have high correlations. This indicates that the learned $\hat G'$ provides some additional information that singleton marginals cannot provide. 

\subsubsection{Scalability to Large Datasets}
\label{sec:expe-scale}
To demonstrate the scalability of \acrshort{ACVAE} compared to \acrshort{CVAE}, we perform an experiment on a larger version of the Citation dataset with $>12$ times more vertices, which \acrshort{CVAE} can not easily scale to due to the cubic time initialization step and the quadratic pairwise marginal evaluations.

We compare the performance of \acrshort{ACVAE} plus the belief propagation refinement on both the empirical Bayes and the saddle-point schemes with the other two baseline methods (\acrshort{VAE} and GraphSAGE). As shown in \Cref{tab:ncrr2}, both schemes of \acrshort{ACVAE} can significantly outperform the baseline methods.\begin{table}[ht]
  \begin{center}
  \caption{NCRR on The Larger Citation Dataset}
\begin{tabular}{ll}
\\
    Method     & NCRR     \\
    \hline
    \acrshort{VAE} &  0.002 \\
    GraphSAGE &	0.002	\\
    \acrshort{ACVAE}$_\textrm{EB+BP}$ & 0.076 \\
    \acrshort{ACVAE}$_\textrm{SP+BP}$ & 0.073	\\
  \end{tabular}
  \end{center}
  \label{tab:ncrr2}
\end{table}

\section{CONCLUSION}
In this paper, we introduce \acrshortpl{ACVAE}, which learn a joint variational distribution on the latent embeddings of input data via optimizing loss function that is a non-uniform average over some tractable correlated ELBOs. To learn the mixture weights, we provide two different options, and compare them on various datasets and tasks. The learned joint variational distribution can be used to perform efficient evaluation using belief propagation. Experiment results show that \acrshortpl{ACVAE} can outperform existing methods for link prediction and hierarchical clustering on three real datasets. Future work will include better understanding the learned graph structures from both options and learning higher-order correlations between latent variables.

\bibliographystyle{plainnat}
\bibliography{paper}

\newpage

\appendix
\section*{Appendix}
In the appendix, we provide more details on our baseline method \acrshort{CVAE} \citep{tang2019correlated} as well as the experiment data pre-processing and protocols.

\section{MORE DETAILS ON \glspl{CVAE}} \label{app:cvae}
\glspl{CVAE} set the prior $p_0^{\textrm{corr}_g}(\bm z)$ to be the uniform average over all of these tractable densities:
 \begin{equation}
p_0^{\textrm{corr}_g}(\bm z)=\frac{1}{|\mathcal A_G|}\sum\limits_{G'=(V,E')\in\mathcal A_G} p_0^{G'}(\bm z),
\end{equation}
where $p_0^{G'}(\bm z)=\prod\limits_{i=1}^np_0(\bm z_i)\prod\limits_{(v_i,v_j)\in E'}\frac{p_0(\bm z_i,\bm z_j)}{p_0(\bm z_i)p_0(\bm z_j)}$ is a prior on a maximal acyclic subgraph $G'=(V,E')$ with the same form as in \Cref{eqn:prior-acyclic}. For each $G' \in \mathcal{A}_G$, we can similarly define a structured variational approximation $q_{\bm\lambda}^{G'}(\bm z | \bm x)$ following the form of \Cref{eqn:prior-acyclic}: 
\begin{equation*}
q_{\bm\lambda}^{G'}(\bm z | \bm x)=\prod\limits_{i=1}^nq_{\bm\lambda}(\bm z_i | \bm x_i)\prod\limits_{(v_i,v_j)\in E'}\frac{q_{\bm\lambda}(\bm z_i,\bm z_j|\bm x_i,\bm x_j)}{q_{\bm\lambda}(\bm z_i|\bm x_i)q_{\bm\lambda}(\bm z_j|\bm x_j)},
\end{equation*}
where $q_{\bm\lambda}(\cdot | \cdot)$ and $q_{\bm\lambda}(\cdot,\cdot | \cdot,\cdot)$ are two conditional density functions that captures the singleton and pairwise variational approximation densities. These two functions need to satisfy the symmetry and consistency properties:
\begin{equation*}
\begin{cases}
q_{\bm\lambda}(\bm z_i, \bm z_j | \bm x_i, \bm x_j)=q_{\bm\lambda}(\bm z_j, \bm z_i | \bm x_j, \bm x_i)\text{\ \ for all }\bm z_i,\bm z_j,\bm x_i, \bm x_j,\\
\int q_{\bm\lambda}(\bm z_i, \bm z_j | \bm x_i, \bm x_j) d\bm z_j=q_{\bm\lambda}(\bm z_i| \bm x_i)\text{\ \ \ \ \ for all }\bm z_i,\bm x_i, \bm x_j.
\end{cases}
\label{eqn:q-constraint}
\end{equation*}

The ELBO in \Cref{eqn:lb1} is an average over potentially exponential many ELBOs. To make computations tractable, \citet{tang2019correlated} simplifies this lower bound and represent it as
\begin{align}
&\mathcal L^{\textrm{CVAE}}(\bm\lambda, \bm\theta):=\sum\limits_{i=1}^n\Big(\mathbb E_{q_{\bm\lambda}(\bm z_i | \bm x_i)}\left[\log p_{\bm\theta}(\bm x_i | \bm z_i) \right] -\text{KL}(q_{\bm\lambda}(\bm z_i | \bm x_i)||p_0(\bm z_i))\Big)-\sum\limits_{(v_i,v_j)\in E}w^{\text{MAS}}_{G, (v_i, v_j)}\cdot\nonumber\\
&\Big(\text{KL}(q_{\bm\lambda}(\bm z_i, \bm z_j | \bm x_i, \bm x_j)||p_0(\bm z_i, \bm z_j))-\text{KL}(q_{\bm\lambda}(\bm z_i | \bm x_i)||p_0(\bm z_i)) - \text{KL}(q_{\bm\lambda}(\bm z_j | \bm x_j)||p_0(\bm z_j))\Big).
\end{align}
Where $w^{\text{MAS}}_{G, e}:=\frac{|\{G'\in\mathcal A_G: e\in G'\}|}{|\mathcal A_G|}$ for each edge $e=(v_i,v_j)$ represents the fraction of $G$'s maximal acyclic subgraphs of $G$ that contain $e$. These weights can be computed easily from the Moore-Penrose inverse of the Laplacian matrix of $G$.

\section{EXPERIMENT DETAILS}\label{app:exp}
\subsection{Dataset Pre-processing Details} \label{app:data}
\paragraph{Epinions}
We follow the same pre-processing scheme as \citet{tang2019correlated}: binarize the rating data and create a bag-of-words binary feature vector for each user. We only retain the items that have been rated for at least 100 times. We construct the graph $G=(V,E)$ and only keep an edge $(v_i, v_j)$ to be in $E$ if both $v_i\rightarrow v_j$ and $v_j\rightarrow v_i$ appear in the original directed graph. At last, we only retain users that have at least one edge in $E$ (i.e. having at least one bi-directional edge in the original dataset).
\paragraph{Citations}
This dataset includes the abstract and the citation information for high-energy physic theory papers on arXiv from 1992 to 2003. We work on all papers from 1998 in this dataset (in total $\approx 2.8\text{K}$ papers). We treat all citation edges as undirected edges and build the graph $G=(V,E)$. We only retain papers that cite or are cited by at least one of the other papers within this subset (for year 1998) of the dataset. We compute the TF-IDF (with stop words removed) for the abstract of each paper as the raw feature vectors, retaining only the coordinates corresponding to the top 50 words. Then we binarize the raw feature vectors that considers only the non-zero entries that are above the median of all of the non-zero entries and use these binarized vectors as the feature vectors.

For the larger experiment on this dataset, we apply the same pre-processing steps but work on the whole dataset (instead of the subset for year 1998).

\paragraph{LibraryThing}
For the link prediction experiment, We follow the same pre-processing scheme as for the Epinions dataset, except that we only retain the items that have been rated for at least 200 times (since this dataset is larger than the Epinions dataset). For the clustering experiment, we follow the same scheme to get a graph $G=(V,E)$, but we do not split the edges to training/testing (since clustering is unsupervised), and apply a normal \acrshort{VAE} to generate the labels (as mentioned in the main paper). This normal \acrshort{VAE} has the same hidden layer size with the one used in testing, but has a smaller latent representation (we use 10) to avoid generating non-reasonable labels due to overfitting.

\subsection{Experimental Protocol} \label{app:exp_proto}
We run 3 runs for each methods for the Epinions experiments,  and 5 runs for the other experiments (except we run only 1 run for the Citation experiments on the larger dataset as in \Cref{sec:expe-scale}). This is since the Epinions experiments work more stable empirically.

For \acrshort{VAE}, \acrshort{CVAE} and \acrshort{ACVAE}, we apply a two-layer feed-forward neural inference network for the singleton variational distribution $q_{\bm\lambda}(\bm z_i | \bm x_i)$'s and a two-layer feed-forward neural generative network for the model distribution $p_{\bm\theta}(\bm x | \bm z)$'s. $q_{\bm\lambda}(\bm z_i | \bm x_i)$ is a diagonal normal distribution with the mean and standard deviation outputted from the inference network and $p_{\bm\theta}(\bm x | \bm z)$ is a multinomial distribution with the logits outputted from the generative networks. The latent dimensionality $d$ is 100 for the Epinions experiments and the LibraryThing clustering, and 10 for the other two link prediction experiments. The hidden layer dimensionality $h_1$ is 300 for the Epinions experiments and 30 for the other experiments. 

For GraphSAGE, we choose to use $K=2$ aggregations, the mean aggregator, and $Q=20$ negative samples to optimize the loss function. The hidden layer size and latent dimensionality we apply to GraphSAGE are the same with that of the standard \acrshort{VAE}.

For \acrshort{CVAE} and \acrshort{ACVAE}, we set the pairwise marginal prior density function to be $p_0(\cdot)=\mathcal N\left(\bm\mu=\bm 0_{2d},\Sigma=\begin{pmatrix}I_d & \tau\cdot I_d\\ \tau\cdot I_d & I_d\end{pmatrix}\right)$ with $\tau=0.99$. For {\sc cvae}$_\textrm{corr}$ and \acrshort{ACVAE}, we model the pairwise variational approximations $q(\bm z_i, \bm z_j | \bm x_i, \bm x_j)$ to be a multi-variate normal distribution that can be factorized across the $d$ dimensions as the product of $d$ independent bi-variate normal distributions. The correlation coefficients of these bi-variate normal distributions are computed from two-layer feed-forward neural networks that taking $\bm x_i$ and $\bm x_j$ as inputs. These two-layer neural networks have latent dimensionality $h_2$ to be 1000 for the Epinions experiments and 100 for the other experiments. For \acrshort{CVAE} and \acrshort{ACVAE} on link prediction experiments, we select the negative sampling parameter $\gamma$ from set of choices, and report the performances with the best average train NCRR metrics. This parameter is selected from $\{0.0001, 0.001, 0.01, 0.1, 1, 10, 100, 1000\}$ for the LibraryThing dataset and the Citation dataset, and $\{0.001, 0.1, 10., 1000.\}$ for the Epinions dataset. For the clustering experiments, we select $\gamma=1$ for \acrshort{CVAE} and \acrshort{ACVAE} since empirically we found this choice gives us a reasonably good performance.

For link prediction, for all methods, we look into the performances for every fixed number of iterations (the specific numbers depend on models) and update the current best test NCRR values if both the train ELBO and the train NCRR reach better values. We report the final current best test NCRR values as the results. For clustering, we update the current best normalized MI scores if the train ELBO reaches better values and report the final current best normalized MI scores.

For \acrshort{ACVAE}, we set the step size parameter (in \Cref{eqn:soft-update}) $\alpha^t=0.1$ to be a constant. We train the parameters using alternating updates as in \Cref{alg:updates}. We switch between updates on the parameters $\bm\lambda$, $\bm\theta$ for an epoch of the edges in $E$, and a single update on the weights $w^{\text{MAS}}_{G, \pi, e}$ according to \Cref{eqn:soft-update}. For the random initialization on the tree weights $w^{\text{MAS}}_{G, \pi, e}$, we just assign random weights to the graph $G=(V,E)$. Then we use Kruskal's algorithm to compute the maximal acyclic subgraph $\tilde G=(V,\tilde E)$ according to these random weights, and set $w^{\text{MAS}}_{G, \pi, e}=I[e\in \tilde E]$. It is straightforward to see that this is a valid initialization for the weights $w^{\text{MAS}}_{G, \pi, e}$'s since these weights relate to the distribution $\bm{\tilde\pi}$ that has all of its mass on the single subgraph $\tilde G$.

For $\acrshort{ACVAE}$, after running the algorithm for some iterations, we use Kruskal's algorithm to compute the maximal acyclic subgraph $\hat G=(V, \hat E)$ on the converged edge weights $\hat w^{\text{MAS}}_{G, \pi, e}$ to find the learned single graph $\hat G'$. This heuristic helps us solve the issues of finding the converged maximal acyclic subgraph if we want to perform an early stopping (recall that we evaluate our metrics for every fixed number of iterations) or if there is an numerical issue.

For all methods, we apply stochastic gradient optimizations and use Adam \cite{kingma2015adam} to adjust the learning rates. We set the step size to be $10^{-3}$. For all methods, we use a batch size $B_1=64$ for sampling the vertices. For \acrshort{CVAE} and \acrshort{ACVAE}, we use a batch size $B_2=256$ for sampling the edges and non-edges.

All experiments are done using Python. The training and evaluations are done with TensorFlow \citep{abadi2016tensorflow} and Numpy. The TF-IDF and the t-SNE embeddings \citep{maaten2008visualizing} in the visualization (\Cref{fig:vis})  are computed using Scikit-learn \citep{pedregosa2011scikit}. For faster computations, we call C++ functions to do belief propagation and the Kruskal's algorithm using Cython \citep{behnel2011cython}.

\end{document}